\documentclass{article}

\usepackage{amsmath}
\usepackage{PRIMEarxiv}

\usepackage[utf8]{inputenc} 
\usepackage[T1]{fontenc}    
\usepackage{hyperref}       
\usepackage{url}            
\usepackage{booktabs}       
\usepackage{amsfonts}       
\usepackage{nicefrac}       
\usepackage{microtype}      
\usepackage{lipsum}
\usepackage{fancyhdr}       
\usepackage{graphicx}       
\usepackage{adjustbox}
\graphicspath{{media/}}     

\usepackage{algorithm}
\usepackage{algorithmic}

\usepackage{dblfloatfix}
\usepackage{multirow}
\usepackage{rotating}
\usepackage{amssymb}
\usepackage{amsmath}
\usepackage{tikz}
\title{A Versatile Multi-View Framework for LiDAR-based 3D Object Detection with Guidance from Panoptic Segmentation}

\author{Hamidreza Fazlali, Yixuan Xu, Yuan Ren, Bingbing Liu\\
Huawei Noah's Ark Lab\\
Toronto, Canada\\
{\tt\small {\{hamidreza.fazlali1@huawei.com, richard.xu2, yuan.ren3, liu.bingbing\}@huawei.com}}
}

\begin{document}
\maketitle

\begin{abstract}
3D object detection using LiDAR data is an indispensable component for autonomous driving systems. Yet, only a few LiDAR-based 3D object detection methods leverage segmentation information to further guide the detection process. In this paper, we propose a novel multi-task framework that jointly performs 3D object detection and panoptic segmentation. In our method, the 3D object detection backbone in Bird's-Eye-View (BEV) plane is augmented by the injection of Range-View (RV) feature maps from the 3D panoptic segmentation backbone. This enables the detection backbone to leverage multi-view information to address the shortcomings of each projection view. Furthermore, foreground semantic information is incorporated to ease the detection task by highlighting the locations of each object class in the feature maps. Finally, a new center density heatmap generated based on the instance-level information further guides the detection backbone by suggesting possible box center locations for objects. Our method works with any BEV-based 3D object detection method, and as shown by extensive experiments on the nuScenes dataset, it provides significant performance gains. Notably, the proposed method based on a single-stage CenterPoint 3D object detection network achieved state-of-the-art performance on nuScenes 3D Detection Benchmark with 67.3 NDS.
\end{abstract}

\section{Introduction}
\label{sec:intro}

Over the past few years, there has been remarkable progress in autonomous vehicles (AVs) vision systems for understanding complex 3D environments \cite{lang2019pointpillars, zhou2018voxelnet, ku2018joint}. 3D object detection is one of the core computer vision tasks that empowers AVs for robust decision-making.
In this task, each of the foreground objects, such as a car, a pedestrian, etc., needs to be accurately classified and localized by a 3D bounding box with 7 degrees of freedom (DOF), including the 3D box center location (x, y, z), size (l, w, h) and yaw angle ($\alpha$). Compared to RGB cameras, LiDAR sensors provide precise 
depth information about the scene, which is valuable for accurate object localization. However, 3D point cloud data provided by the LiDAR scanners are sparse, irregular and unordered, which make the exploitation of such information difficult. 

\begin{figure}[t]
  \centering
  \includegraphics[width=0.6\linewidth]{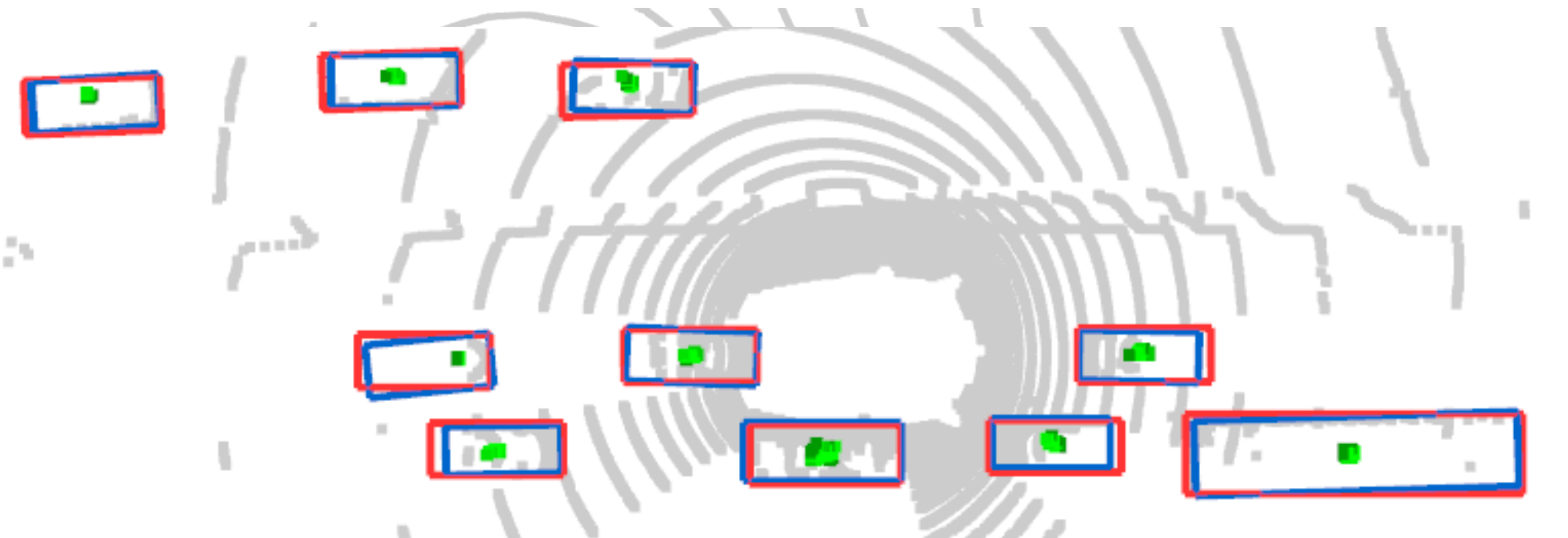}
   \caption{The proposed center density heatmap (colored in \textcolor{green}{green}) guides the detection head toward possible box center regions on the BEV plane. The \textcolor{blue}{blue} bounding boxes represent predictions, while the \textcolor{red}{red} bounding boxes represent ground truth. Best viewed in color.}
   \label{fig:figure1}
\end{figure}

LiDAR-based 3D object detection methods rely on different strategies for 3D point cloud data representation. Some of these detection methods \cite{qi2018frustum, shi2019pointrcnn, yang2019std} are categorized as point-based methods. These methods directly process the raw point cloud to extract useful information. While they usually achieve high accuracy, their computational cost is significant due to the large number of points in each LiDAR scan. The other subset of LiDAR-based 3D object detection methods are known as grid-based methods \cite{zhou2018voxelnet, yan2018second, lang2019pointpillars, shi2020points}. These methods transform the unordered point cloud into regular 3D volumetric grids, \textit{i.e.}, voxels or pillars, and extract discriminative features from the points inside each gird cell. The extracted features are further processed by 2D or 3D Convolutional Neural Networks (CNN). 
Although point sub-sampling helps grid-based methods to be computationally efficient,
some information is lost during projection and discretization \cite{yang2018pixor}.

In each LiDAR scan, there is a vast number of points to be processed. Many of these points represent the background region, including drivable surface, sidewalk, vegetation, etc. Feeding all this information to a 3D object detection algorithm without providing extra clues, such as semantic information, makes the recognition and localization process challenging. Several works \cite{shi2019pointrcnn, yang2019std, sun2021rsn} have exploited 
a binary or semantic segmentation model for filtering background points or providing extra semantic features that can guide the proposal generation or detection process. 

Inspired by this notion of providing guidance, we resort to 3D panoptic segmentation as an auxiliary task for guiding and further improving the performance of Bird's-Eye-View (BEV) based 3D object detection algorithms. A 3D panoptic segmentation method predicts the semantic class label and performs instance-level segmentation for each point in the 3D space, both of which are useful as guidance signals for detecting objects. In addition, guiding a BEV-based detection model with features learned from a Range-View (RV) based network can reduce the sparsity of features representation in BEV projection. Here, we validate these ideas by training a BEV-based 3D object detection method in conjunction with an RV-based 3D panoptic segmentation method. More specifically, we supplement the BEV-based detection backbone with additional RV features extracted from the panoptic segmentation backbone, providing a rich set of multi-view information to aid the detection. Moreover, we exploit the semantic labels of the foreground objects estimated by the panoptic segmentation network to refine the 3D object detection backbone features. Finally, a center density heatmap in the BEV plane is designed based on the instance-level information obtained from the panoptic segmentation, highlighting regions that contain box centers of objects.
In conjunction, the augmented backbone features, foreground semantic labels, and the center density heatmap guide the detection head towards a more accurate 3D box recognition and localization. We will describe our multi-task framework based on the single-stage CenterPoint 3D object detection method \cite{yin2021center}, but later in the experimental results section, we will quantitatively demonstrate that our approach can help any existing BEV-based 3D object detection method.

Our contributions can be summarized into four-fold. (1) We propose a multi-task framework that jointly learns 3D panoptic segmentation and 3D object detection for improving the 3D object recognition and localization. To the best of our knowledge, this is the first framework that leverages both the semantic- and instance-level information concurrently for improving 3D object detection. (2) The framework is also designed to
be easily attached to any BEV-based object detection method as a plug-and-play solution to boost its detection performance. (3) With extensive experiments conducted on the nuScenes dataset \cite{caesar2020nuscenes},  which includes both the panoptic and 3D box information, we validate the effectiveness of our method with different BEV-based 3D object detection methods. (4) We conduct ablation studies to further examine the usefulness of each add component 
for performance improvement.

\section{Related Work}

\label{sec:relwork}
\textbf{3D Object Detection with Point-based Methods}. The point-based methods take in the unordered sets of 3D point cloud and rely on PointNet \cite{qi2017pointnet} or PointNet++ \cite{qi2017pointnet++} for feature extraction. FPointNet \cite{qi2018frustum} uses the 2D object proposals from camera images to filter the point cloud and then uses PointNet for 3D object detection based on the proposal regions. Its performance suffers as proposal regions generated from RGB images lack accurate 3D information. PointRCNN \cite{shi2019pointrcnn} addressed this problem by first segmenting the foreground points using PointNet++ and then refining the proposals using the segmentation features. STD \cite{yang2019std} uses PointNet++ for proposal generation and then further densifies the point features within each proposal using a pooling strategy. Generally, the point-based methods have a larger receptive field compared to the grid-based methods; however, their computational complexity is very high due to the vast number of points in each outdoor LiDAR scan \cite{shi2020pv}.

\textbf{3D Object Detection with Grid-based Methods}. These methods divide the 3D space into volumetric grids known as voxels, so they can be processed by 2D or 3D CNN. Earlier methods encode each voxel with some hand-crafted features. PIXOR \cite{yang2018pixor} encoded each voxel based on the occupancy and reflectance of points. Complex-YOLO \cite{simony2018complex} encoded each grid cell with the maximum height, maximum intensity, and normalized point density. In order to extract a more useful and richer set of features, VoxelNet \cite{zhou2018voxelnet} designed a Voxel Feature Encoder (VFE) layer to leverage the power of deep learning for voxel feature learning and then used a 3D CNN for its detection backbone. PointPillars \cite{lang2019pointpillars} reduced the number of voxels to one along the height dimension, improving both the inference time and detection accuracy. In \cite{bewley2020range}, 3D object detection was done in RV by using a range-conditioned dilation (RCD) layer that addresses the problem of object scale change in the range image. To take advantage of both BEV and RV point cloud representations, CVCNet \cite{chen2020every} used Hybrid-Cylindrical-Spherical (HCS) voxels. RSN \cite{sun2021rsn} was another multi-view fusion method that first performed binary segmentation on the RV and then applied sparse convolutions on the foreground voxels with the learned RV features to detect objects. To address the imbalance of voxel sparsity between object classes,  \cite{chen2020object} considers a limited number of non-empty object voxels as hotspots and the detection head predicts these hot spots and the corresponding boxes.
Following the success of CenterNet \cite{duan2019centernet} in 2D object detection, CenterPoint \cite{yin2021center} used an anchor-free 3D object detection head (center-head) that first detects the center point of each object in the BEV plane and subsequently regresses the bounding box dimensions. 

\section{Preliminaries}
\label{sec:prl}
\textbf{Cluster-free 3D Panoptic Segmentation (CPSeg).} CPSeg \cite{li2021cpseg} uses a dual-decoder architecture to conduct panoptic segmentation without generating object proposals or using clustering algorithms. The backbone takes in the RV representation of the LiDAR point cloud and provides multi-scale feature maps. The semantic decoder provides semantic labels, while the instance decoder predicts the object mass centroid as instance embedding for each point. Different from the semantic decoder, which only utilizes encoded feature maps, the instance decoder benefits from the additional information of computed surface normals. Then, the cluster-free instance segmentation head dynamically groups points with similar instance embedding as pillars in BEV, and objects are segmented by calculating the connectivity between pillars through a pairwise embedding comparison matrix.


\textbf{CenterPoint 3D Object Detection.} Originally, CenterPoint is a two-stage 3D object detection method, where bounding boxes are regressed in the first stage and further refined in the second stage. Specifically, the second stage takes in the backbone feature maps in the BEV plane and considers information at locations where four sides of the first-stage bounding box are located.
During training, the model estimates
a center heatmap for each object class and 
other regression targets, including the box center offset, size, angle, and velocity. In the estimated center heatmap for each class, the local-maxima values represent the confidence scores. Their locations on the map are used to estimate the other regression targets. Overall, CenterPoint removes the need for anchor boxes in 3D object detection, which were inherited from the 2D object detection and were challenging to fit in the 3D space.

\section{Proposed Approach}

\label{sec:proposed}
\begin{figure*}[t]
  \centering
  \includegraphics[width=1\linewidth]{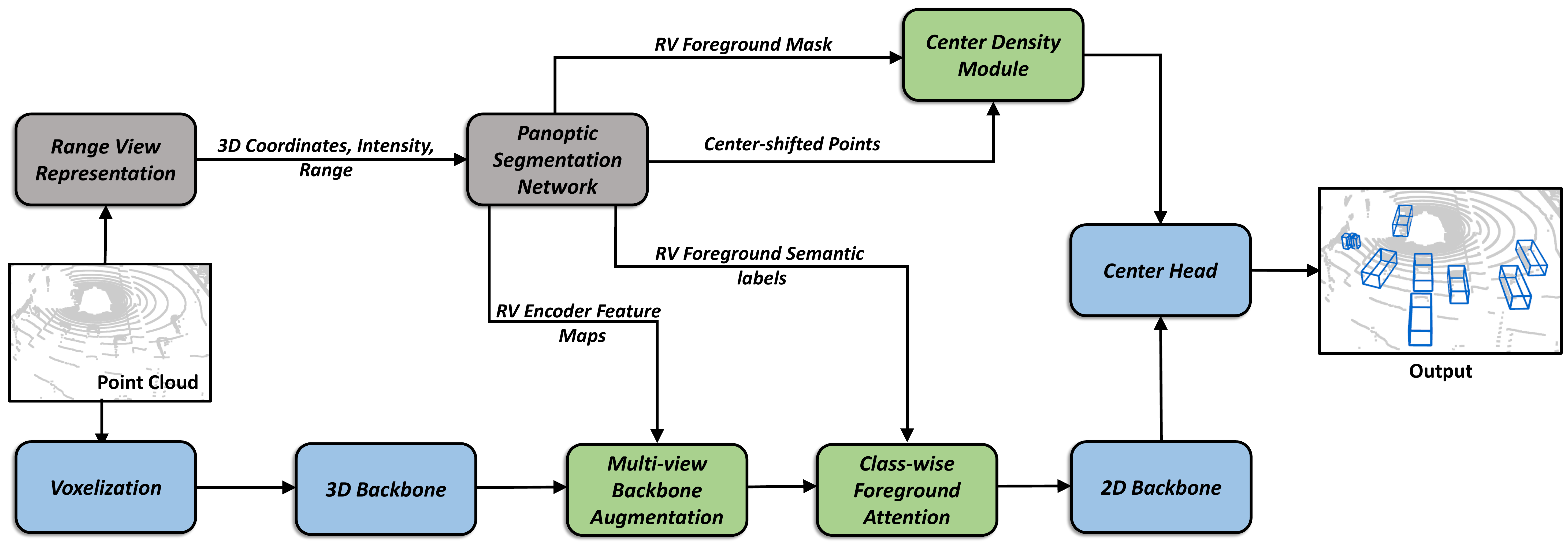}
   \caption{Block diagram of the proposed method. The \textcolor{gray}{\textbf{gray}} blocks represent the CPSeg model \cite{li2021cpseg} and the \textcolor{blue}{\textbf{blue}} blocks represent the single-stage CenterPoint model \cite{yin2021center}. The \textcolor{green}{\textbf{green}} blocks are the proposed modules for combining the 3D panoptic segmentation and 3D object detection under a multi-task framework. Best viewed in color.}
   \label{fig:figure2}
\end{figure*}

The block diagram of the proposed multi-task framework is shown in Figure \ref{fig:figure2}. This method receives raw LiDAR point cloud data as input and outputs both the panoptic segmentation and 3D object detection results. For panoptic segmentation, we resort to CPSeg \cite{li2021cpseg} for its state-of-the-art and real-time performance. We made some architectural modifications to the CPSeg model for speeding up the proposed multi-task framework. The details of these changes are described in the supplementary materials. Furthermore, instead of predicting mass-center offsets as in the original CPSeg, the instance segmentation head in our modified CPSeg provides the 3D box-center offsets, useful for guiding the 3D object detection. Other than center offsets, CPSeg also provides its encoder feature maps and generated foreground semantic predictions to aid the detection model, as shown in Figure \ref{fig:figure2}.

For 3D object detection, we chose CenterPoint for its performance superiority compared to the anchor-based 3D object detection methods. We use the single-stage CenterPoint method based on VoxelNet backbone \cite{duan2019centernet}. The CenterPoint consists of two main components: the backbone and the detection head. As shown in Figure \ref{fig:figure2}, the detection backbone consists of the voxelization module that divides a point cloud into volumetric voxel grids, the 3D CNN backbone that learns 3D structural features, and a 2D CNN backbone that further processes the learned features in BEV. The detection head includes a group of class-specific center heads that predict center heatmap and other bounding box regression targets. We remove the second stage refinement process within the CenterPoint in our multi-task framework as information from panoptic segmentation is found to be sufficient in guiding the detection head to accurate detections.
Moreover, in the Experiments section, we will show that our multi-task framework can easily work with any existing BEV-based 3D object detection method by simply replacing the detection backbone and head. 

The 3D object detection and panoptic segmentation methods in our framework are trained jointly. First, the backbone of the detection network is augmented with the addition of a rich set of RV features from the panoptic segmentation encoder. 
Furthermore, foreground semantic labels and instance box center offsets estimated by the panoptic segmentation network are also injected to guide the 2D backbone and detection head to attend to the locations of potential objects and their centers, respectively. As illustrated in Figure \ref{fig:figure2}, the integration takes place in the multi-view backbone augmentation, class-wise foreground attention, and center density heatmap modules. More details about these blocks are covered in the following subsections.
\subsection{Multi-View Backbone Augmentation}

RV and BEV representations of point clouds enable the design of efficient 3D perception models. Feature extractors of state-of-the-art panoptic segmentation models commonly rely on range view \cite{milioto2020lidar, sirohi2021efficientlps}, while most well-known 3D detection methods operate on the BEV plane \cite{yan2018second, lang2019pointpillars, yin2021center, chen2020object}. However, each form of projection has its strengths and weaknesses. For example, features representation in RV are denser and align with the LiDAR scan patterns. Thus small objects, such as pedestrians, traffic cones, motorcycles, and bicycles, are more visible and easier to be detected and classified. However, determining the sizes and boundaries of crowded or distant objects is difficult in RV due to occlusions and size variations. On the other hand, BEV avoids the issues presented in RV, but its sparse and coarse representations make it challenging to detect smaller objects. 

Similar to \cite{chen2017multi, chen2020every}, we attempt to leverage the strength of each view to improve the performance of 3D object detection.
Here, we introduce the concept of feature weighting to combine multi-view features, by adaptively weighting each feature value in RV and BEV maps based on its perceived importance in boosting the detection performance.
The multi-view backbone augmentation is composed of two steps. 
First, RV feature maps from the segmentation model backbone are projected to the BEV plane and further down-sampled to match the resolution of the extracted BEV feature maps in the detection backbone. Then, the RV- and BEV-based feature maps are fused using a proposed space-channel attention module, which weights each feature map based on its usefulness for the detection task. These two steps are elaborated below.

\begin{figure}[h]
  \centering
  \includegraphics[width=1\linewidth]{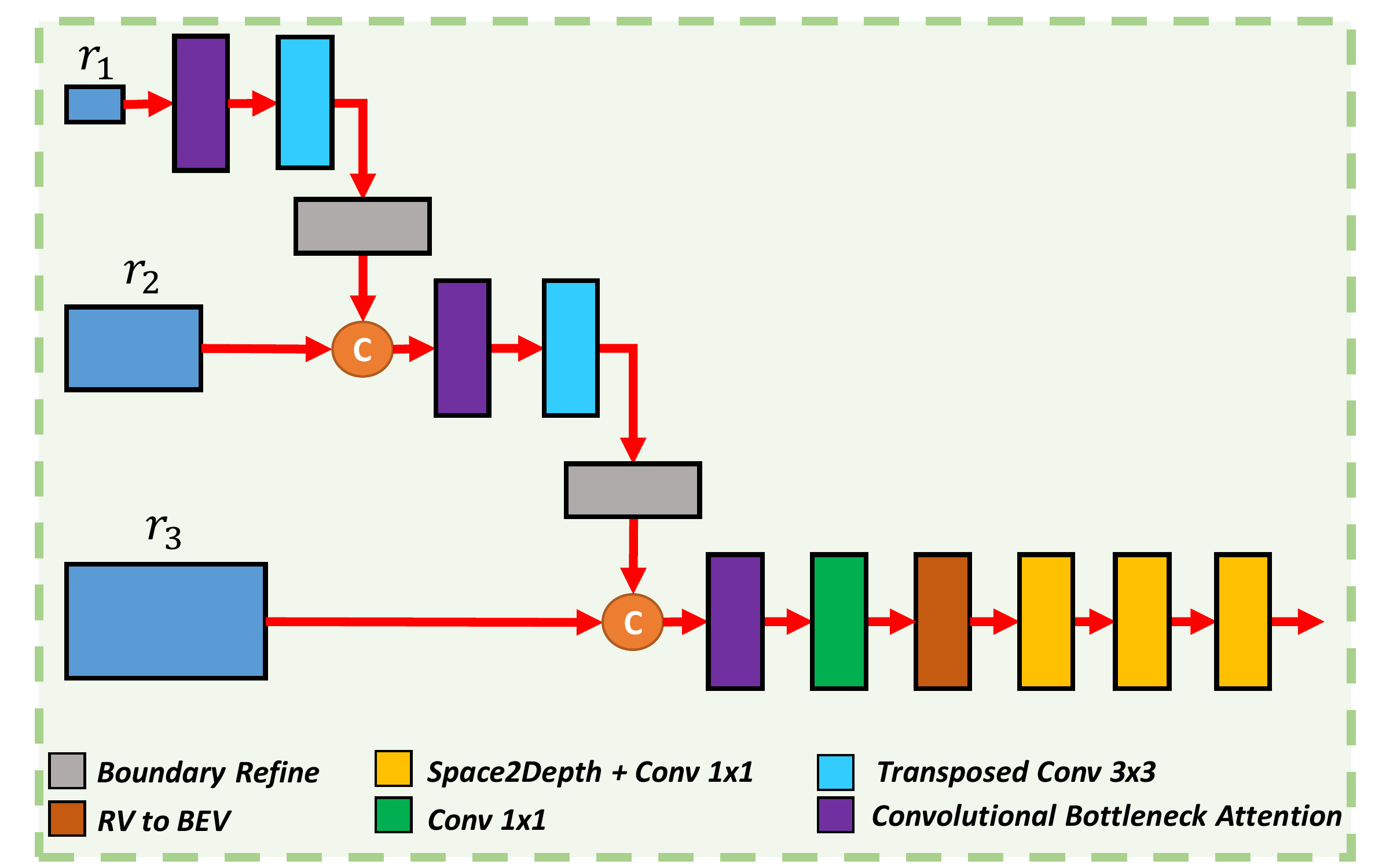}
   \caption{Cascade feature fusion module. The selected RV feature maps are fused and the projected to BEV plane and then down-sampled to match the detection backbone feature map resolution. Best viewed in color.}
   \label{fig:figure3}
\end{figure}

\subsubsection{Cascade RV Feature Fusion Module}

The RV feature maps generated in the panoptic segmentation backbone can help augment the detection backbone, which operates in the BEV plane, in detecting smaller objects that are otherwise not properly represented.

A cascade feature fusion module, as shown in Figure \ref{fig:figure3}, processes multi-scale RV feature maps and prepares them for the detection backbone augmentation. The coarser RV feature maps, $r_{1}$ and $r_{2}$, are obtained from intermediate layers of the CPSeg encoder, which contain contextual RV information that can benefit multiple tasks. On the other hand, high-resolution feature maps, $r_{3}$, encode additional geometric information of point cloud in the RV plane to emphasize the locations and presence of objects. More specifically, the learned geometric features extracted from surface normal vectors originated in CPSeg are concatenated with the 3D Cartesian coordinates associated with each point. 

In the proposed cascade feature fusion module, starting from the coarsest-scale, feature maps $r_{1}$ are first processed by a Convolutional Bottleneck Attention Module (CBAM) \cite{woo2018cbam}. This module is responsible for adaptive feature refinement along space- and channel-dimensions. Then, the resulted feature maps are up-sampled by a factor of 2 using a $3 \times 3$ Transposed Convolution layer and passed to the boundary refinement layer to reduce the up-sampling artifacts. These up-sampled features are then concatenated to higher-resolution feature maps, $r_{2}$, and the same previously mentioned operations are applied on the concatenated feature maps. 
After the features are concatenated at the highest-resolution scale with $r_{3}$, they are passed through a CBAM and a $1 \times 1$ convolution layer, and subsequently projected to the BEV plane. Finally, a sequence of down-sampling blocks reduces the resolution of the features to the specifications of the detection backbone. Each $Space2Depth$ operation reduces the spatial resolution of the feature maps by half and doubles the channel number, and the accompanied $1 \times 1$ convolution layer compresses the feature maps along the depth dimension. 

\begin{figure}[t]
  \centering
  \includegraphics[width=1\linewidth]{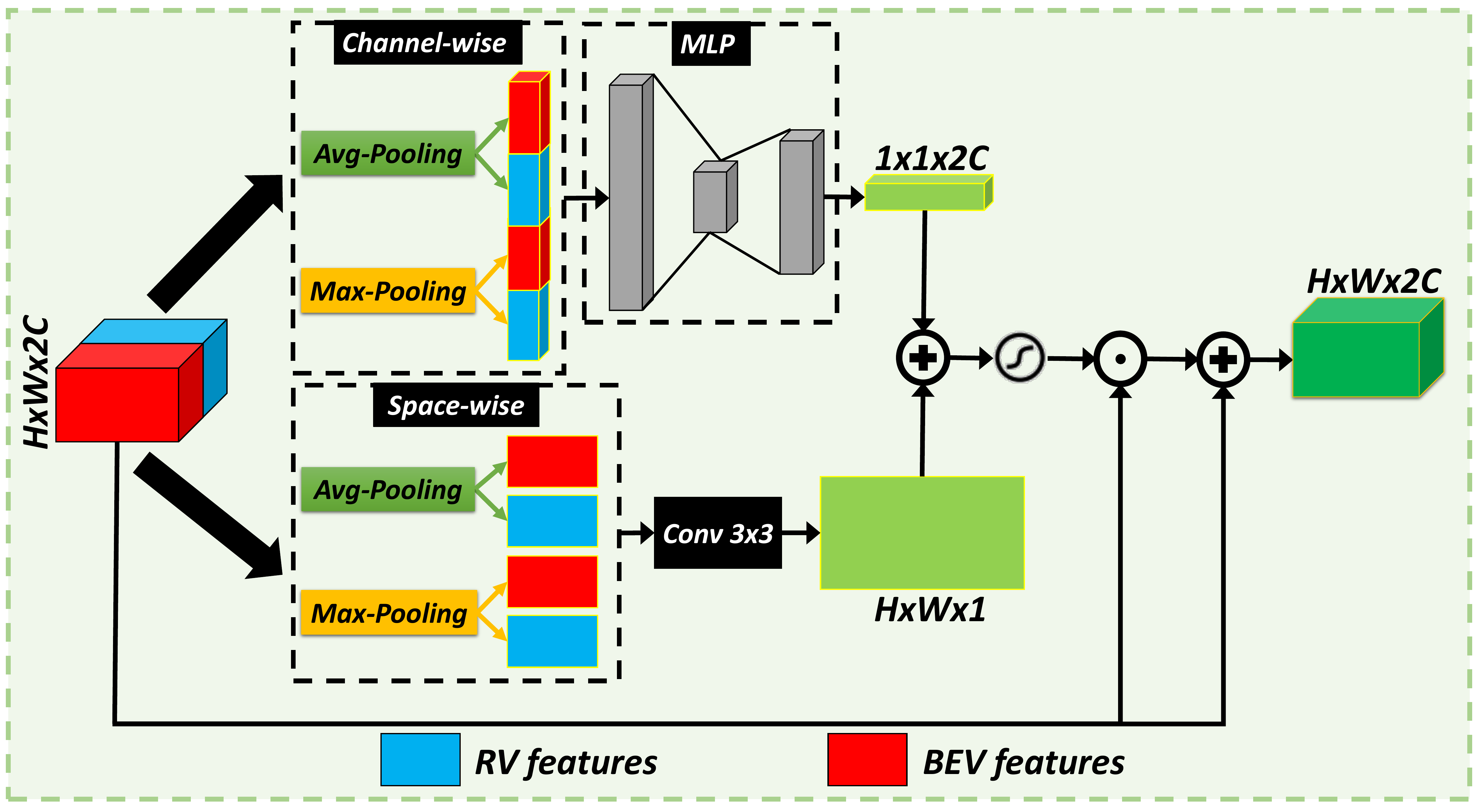}
   \caption{Attention-based RV-BEV feature weighting module. The model learns to highlight the best feature values among the RV and BEV representations along space and channel dimensions. Best viewed in color.}
   \label{fig:figure4}
\end{figure}

\begin{figure}[t]
  \centering
  \includegraphics[width=1\linewidth]{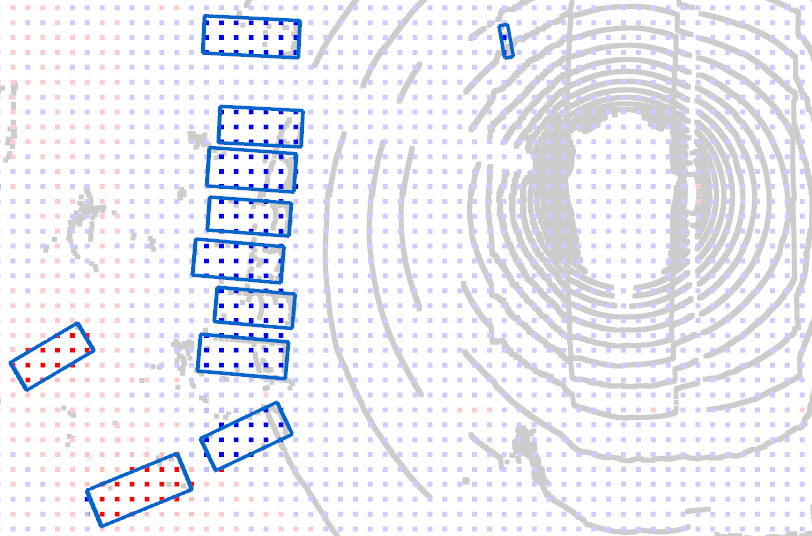}
   \caption{Visualization of the attention map generated by the attention-based RV-BEV feature weighting module. Regions in \textcolor{red}{red} and \textcolor{blue}{blue} indicate locations where BEV and RV features are deemed more significant with higher weights, respectively. Boxes in \textcolor{blue}{blue} are ground truth object bounding boxes. Colors in background are shaded for better visualization. Best viewed in color.}
   \label{fig:figure_richard}
\end{figure}

\subsubsection{Attention-based RV-BEV Feature Weighting Module}
After the cascade RV feature fusion module, feature maps extracted from two different views are concatenated. An attention-based weighting module is crucial so that the model learns to adaptively focus on feature values from a particular view given the scenario.
To this end, a modified CBAM \cite{woo2018cbam} is proposed here to provide useful weighting for RV and BEV features. 

The diagram of this module is shown in Figure \ref{fig:figure4}. The attention-based RV-BEV weighting module takes in the concatenated feature maps from both views, and distributes them to channel- and space-attention streams. The channel-attention stream applies max-pooling and average-pooling across \textit{x}- and \textit{y}-axis of the BEV and RV feature maps, resulting in 4 sets of feature vectors, which are then fed to a multi-layer perceptron (MLP) to learn an attention value for each channel in RV and BEV feature maps. The space-attention stream applies the same pooling operations depth-wise to obtain 4 feature maps, which are then fed to a $3 \times 3$ convolution layer with a dilation rate of 3 to learn spatial attention values. The resulted feature map from the space-attention stream and the feature vector from the channel-attention stream are both broadcasted to the same shape, summed, and passed through a $sigmoid$ activation function to generate an attention map. By applying this attention map to the input feature maps, only the features values from either views that help the detection task are highlighted, and the other feature values are suppressed. This enables the detection backbone to dynamically choose features from a particular view given the presented scene. For instance, as shown in Figure \ref{fig:figure_richard}, the attention-based RV-BEV weighting module assigns higher weights for the RV features representing nearby and smaller objects, while for the occluded and distant objects it tends to favor BEV features more.

\subsection{Class-wise Foreground Attention Module}
In the proposed multi-task framework, the panoptic segmentation estimates the semantic labels for all points in the RV plane, covering both foreground and background objects. However, for 3D object detection, we are only interested in detecting foreground objects. Thus, leveraging the foreground semantic information can ease the detection task by placing more focus on class-specific foreground regions. 

For this purpose, a class-wise foreground attention module is designed, as shown in Figure \ref{fig:figure5}. The module takes in the estimated probability maps for each foreground object category from CPSeg. These maps are projected to the BEV plane and down-sampled via max-pooling to match the resolution of the combined RV-BEV feature maps. For each foreground object category, a class-wise attention branch is created, which performs element-wise multiplication between the probability map of the specific class and the input feature maps, and then compresses the channel depth of the resulted output through a $1 \times 1$ convolution layer. After the attention branch, feature maps are gathered and passed through another $1 \times 1$ convolution layer to keep the channel size the same as the input feature map. Finally, input feature maps are added to these semantically rich feature maps via a skip connection. Overall, this module embeds the foreground semantic information in the feature maps, which can help both the classification and localization tasks.

\subsection{Center Density Heatmap Module} 
The aim of this module is to provide instance-level information for the detection head. The estimated 3D boxes center offsets for each point from CPSeg are used to create a heatmap of potential locations of 3D boxes centers to guide the detection head.
More specifically, an estimated foreground mask from CPSeg is used to filter out background points, as the center offsets from those points are not meaningful. The remaining foreground points are shifted according to the offset predictions and projected to the BEV plane. Then, the center density heatmap is generated as,
\begin{equation}
  H(x,y) = Tanh(log(C(x,y)+1))
  \label{eq:eq1}
\end{equation}

\noindent where $C(x,y)$ is the number of projected points, $H(x,y)$ is the resulted center density heatmap, and $x$ and $y$ represent the horizontal and vertical coordinates on the BEV plane, respectively. 
The $Tanh$ activation function is applied after a $log$ operation to constrain the heatmap values in $[0, 1]$.

This gray-scale center density heatmap is applied to the feature maps extracted from the 2D detection backbone before using used by the detection head. An element-wise multiplication is performed between feature maps and the estimated center density heatmap and the result is added to the feature maps.
As shown in Figure \ref{fig:figure1}, the center density heatmap provides an effective guide to direct the detection head towards each possible box center region.


\section{Experiments}

\label{sec:exp}
In this section, we introduce the dataset used (Sec. \ref{subsec:dataset}) and the implementation details of our framework and the methods (Sec. \ref{subsec:impl}). Moreover, we compare the proposed method with other state-of-the-art methods (Sec. \ref{subsec:results}) and conduct ablation studies to show the effectiveness of each new component in our framework (Sec. \ref{subsec:abl}).

\subsection{Dataset}
\label{subsec:dataset}
The nuScenes \cite{caesar2020nuscenes} is a popular large-scale driving-scene dataset. It provides both ground-truth 3D boxes and panoptic labels. This is in contrast to the other object detection datasets such as KITTI \cite{geiger2012we} and Waymo Open Dataset \cite{sun2020scalability}, which only provide the 3D ground-truth boxes. Moreover, as the proposed method relies on both ground-truth panoptic labels and the 3D boxes, we used the nuScenes dataset for both training and evaluation.

The nuScenes dataset contains 1000 sequences of driving scenes. Overall, 40K frames are annotated for the 3D object detection task with 10 object categories, from which 28K, 6K, and 6K frames are for training, validation, and test sets, respectively. In this dataset, the mean Average Precision (mAP) is one of the metrics used for 3D object detection, which is calculated on the BEV plane based on different center distance thresholds \textit{i.e.} 0.5m, 1.0m, 2.0m, and 4.0m. Another important metric used is the nuScenes Detection Score (NDS), which is a weighted sum of mAP and 4 other metrics that determine the quality of the detections in terms of box locations, sizes, orientations, attributes, and velocities \cite{caesar2020nuscenes}.

\subsection{Implementation Details}
\label{subsec:impl}
We used the Pytorch deep learning library \cite{paszke2019pytorch} and based our implementation on the OpenPCDet \cite{openpcdet2020}, an open-source project for LiDAR-based 3D object detection. In order to demonstrate that the proposed framework can improve any BEV-based 3D object detection method, we used SECOND \cite{yan2018second} and PointPillars \cite{lang2019pointpillars} 3D object detectors in our framework as alternatives to the CenterPoint \cite{yin2021center}. For each experiment using our multi-task framework, the panoptic segmentation model CPSeg and one of the 3D object detection models, such as CenterPoint, are trained jointly from scratch in an end-to-end manner.


To supervise the detection model, the focal loss is used for regressing the center heatmap and for all the other regression targets, the Smooth L1 loss is exploited. The focal loss is computed over all locations on the output heatmap, while the regression loss is only calculated over positive locations. To supervise the panoptic segmentation output, we followed CPSeg for all the loss terms. All the models were trained for 120 epochs on 8 Tesla V100 GPUs with Adam optimizer and a weight decay of $10^{-2}$. Finally, the learning rate was set to $10^{-3}$ and the One Cycle policy was used for learning rate scheduling. 


\subsection{Results}
\label{subsec:results}

\begin{table*}[t]
\scalebox{0.92}{
  \centering
  \begin{tabular}{*{13}{c}}
    \toprule
    \textbf{Method} & \textbf{mAP} & \textbf{NDS} & \textbf{Car} & \textbf{Truck} & \textbf{Bus} & \textbf{Trailer} & \textbf{CV} & \textbf{Ped} & \textbf{Motor} & \textbf{Bic} & \textbf{TC} & \textbf{Barrier} \\
    \midrule
    CenterPoint \cite{yin2021center} & 56.4 & 64.8 & 84.7 & 54.8 & 67.2 & 35.3 & 17.1 & 82.9 & 57.4 & 35.9 & 63.3 & 65.1 \\
    Ours & \textbf{60.3} & \textbf{67.1} & \textbf{85.1} & \textbf{57.1} & \textbf{68.3} & \textbf{43.6} & \textbf{20.5} & \textbf{84.7} & \textbf{62.5} & \textbf{43.6} & \textbf{71.5} & \textbf{66.0} \\
    \textit{Improvement} & \textit{+3.9} & \textit{+2.3} & \textit{+0.4} & \textit{+2.3} & \textit{+1.1} & \textit{+8.3} & \textit{+3.4} & \textit{+1.8} & \textit{+5.7} & \textit{+7.7} & \textit{+8.2} & \textit{+0.9} \\
    \bottomrule
  \end{tabular}}
  \caption{Comparison of the proposed method with the two-stage CenterPoint method on the nuScenes validation set. In the columns, CV, Ped, Motor, Bic, and TC are abbreviations for Construction Vehicle, Pedestrian, Motorcycle, Bicycle, and Traffic Cone, respectively.}
  \label{tab:valset}
\end{table*}

\begin{table*}[t]
  \centering
  \scalebox{0.92}{
  \begin{tabular}{*{13}{c}}
    \toprule
    \textbf{Method} & \textbf{mAP} & \textbf{NDS} & \textbf{Car} & \textbf{Truck} & \textbf{Bus} & \textbf{Trailer} & \textbf{CV} & \textbf{Ped} & \textbf{Motor} & \textbf{Bic} & \textbf{TC} & \textbf{Barrier} \\
    \midrule
    WYSIWYG \cite{hu2020you} & 35.0 & 41.9 & 79.1 & 30.4 & 46.6 & 40.1 & 7.1 & 65 & 18.2 & 0.1 & 28.8 & 34.7 \\
    PointPillars \cite{lang2019pointpillars} & 30.5 & 45.3 & 68.4 & 23.0 & 28.2 & 23.4 & 4.1 & 59.7 & 27.4 & 1.1 & 30.8 & 38.9 \\
    PointPainting \cite{vora2020pointpainting} & 46.4 & 58.1 & 77.9 & 35.8 & 36.2 & 37.3 & 15.8 & 73.3 & 41.5 & 24.1 & 62.4 & 60.2\\
    PMPNet \cite{yin2020lidar} & 45.4 & 53.1 & 79.7 & 33.6 & 47.1 & 43.1 & 18.1 & 76.5 & 40.7 & 7.9 & 58.8 & 48.8\\
    SSN \cite{zhu2020ssn} & 46.4 & 58.1 & 80.7 & 37.5 & 39.9 & 43.9 & 14.6 & 72.3 & 43.7 & 20.1 & 54.2 & 56.3\\
    CBGS \cite{zhu2019class} & 52.8 & 63.3 & 81.1 & 48.5 & 54.9 & 42.9 & 10.5 & 80.1 & 51.5 & 22.3 & 70.9 & 65.7\\
    CVCNet \cite{chen2020every} & 55.3 & 64.4 & 82.7 & 46.1 & 46.6 & 49.4 & 22.6 & 79.8 & 59.1 & 31.4 & 65.6 & 69.6\\
    CenterPoint \cite{yin2021center} & 58.0 & 65.5 & \textbf{84.6} & \textbf{51.0} & 60.2 & 53.2 & 17.5 & 83.4 & 53.7 & 28.7 & 76.7 & 70.9\\
    HotSpotNet \cite{chen2020object} & 59.3 & 66.0 & 83.1 & 50.9 & 56.4 & 53.3 & 23.0 & 81.3 & 63.5 & 36.6 & 73.0 & \textbf{71.6}\\
    \bottomrule
    Ours & \textbf{60.9} & \textbf{67.3} & \textbf{84.6} & 50.0 & \textbf{63.2} & \textbf{55.3} & \textbf{23.4} & \textbf{83.7} & \textbf{65.1} & \textbf{38.9} & \textbf{76.8} & 68.2\\
    \textit{Improvement} & \textit{+1.6} & \textit{+1.3} & \textit{0.0} & \textit{-1.0} & \textit{+3.0} & \textit{+2.0} & \textit{+0.4} & \textit{+0.3} & \textit{+1.6} & \textit{+2.3} & \textit{+0.1} & \textit{-3.4}\\
    \bottomrule
  \end{tabular}}
  \caption{Comparison of the state-of-the-art methods on the nuScenes test set. In the columns, CV, Ped, Motor, Bic and TC represent Construction Vehicle, Pedestrian, Motorcycle, Bicycle, and Traffic Cone, respectively.}
  \label{tab:testset}
\end{table*}

First, we present the detection results of our method and the CenterPoint \cite{yin2021center} on the nuScenes validation set, as shown in Table \ref{tab:valset}. It can be seen that the proposed multi-task framework, which is based on the single-stage CenterPoint detection model, outperforms the original two-stage CenterPoint considerably in NDS and mAP. More specifically, the proposed method significantly improves the AP for all the object categories, specially the smaller ones, such as motorcycles, bicycles, and traffic cones. This performance boost is made possible by exploiting the panoptic segmentation, particularly the RV features in which the smaller objects are better preserved and represented. Qualitative results on two sample LiDAR frames of the validation set are shown in Figure \ref{fig:figure7}, while additional quantitative and qualitative comparisons are included in the supplementary materials.

The comparison results with the state-of-the-art methods on the nuScenes test set are shown in Table \ref{tab:testset}. It can be seen that the proposed method surpasses others by a considerable margin, increasing both the NDS and mAP by $1.3\%$ and $1.6\%$, respectively. This improvement is even higher compared to the original two-stage CenterPoint model, with $1.8\%$ and $2.9\%$ increases in NDS and mAP, respectively. 

\begin{figure*}[!t]
  \centering
  \includegraphics[width=1\linewidth]{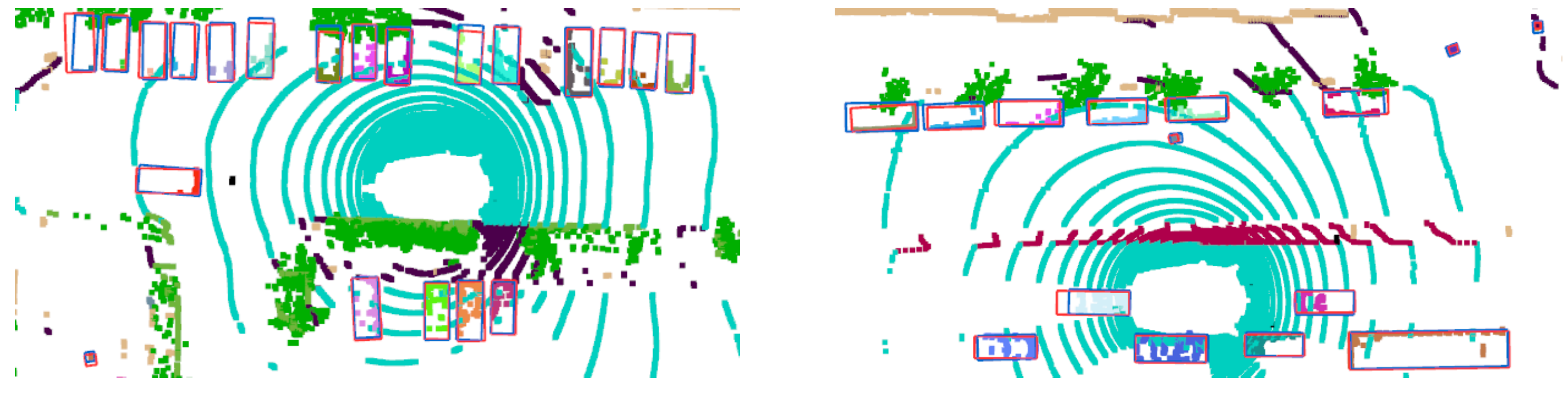}
   \caption{Examples of qualitative results, containing the predicted bounding boxes (in \textcolor{blue}{blue}), ground truth bounding boxes (in \textcolor{red}{red}), and panoptic segmentation results. Best viewed in color.}
   \label{fig:figure7}
\end{figure*}

\begin{table}
  \centering
  \begin{tabular}{*{5}{c}}
    \toprule
    \textbf{Method} & \textbf{MBA} & \textbf{CFA} & \textbf{CDH} & \textbf{NDS} \\
    \midrule
    Baseline &  &  &  & 63.8  \\
     & \checkmark &  &  & 66.5 \\
     & \checkmark & \checkmark &  & 66.9  \\
     & \checkmark & \checkmark & \checkmark & \textbf{67.1} \\
    \bottomrule
  \end{tabular}
  \vspace{5 pt}
  \caption{Effects of different proposed components on the performance improvement evaluated on the nuScenes validation set. In the columns MBA, CFA, and CDH  are abbreviations for multi-View backbone augmentation, Class-wise foreground attention, center density heatmap modules, respectively. The baseline is the single-stage CenterPoint method with VoxelNet (VN) backbone.}
  \label{tab:ablations}
\end{table}

\subsection{Ablation Studies}
\label{subsec:abl}
\textbf{Effects of each proposed component}
The key components proposed in our multi-task framework include the multi-view backbone augmentation, class-wise foreground attention, and center density heatmap modules. The effect of each of these components on the performance of the proposed method evaluated using the nuScenes validation set is shown in Table \ref{tab:ablations}. Based on these results, the multi-view backbone has the most impact on performance improvement by providing RV feature maps to augment the detection backbone. Moreover, the use of class-wise foreground attention and center density heatmap modules also contribute to the performance gains in detection scores considerably. This suggests that the injection of panoptic segmentation information provides helpful guidance for CenterPoint instead of creating confusion.

\begin{table}[t]
  \centering
  \begin{tabular}{*{3}{c}}
    \toprule
    \textbf{Method} & \textbf{mAP} & \textbf{NDS} \\
    \midrule
    PointPillars \cite{lang2019pointpillars} & 43.0 & 56.8 \\
    Multi-Task+PointPillars & \textbf{50.5} & \textbf{60.5} \\
    \textit{Improvement} & \textit{+7.5} & \textit{+3.7} \\
    \midrule
    SECOND \cite{yan2018second} & 51.7 & 62.6 \\
    Multi-Task+SECOND & \textbf{56.2} & \textbf{64.8} \\
    \textit{Improvement} & \textit{+4.5} & \textit{+2.2} \\
    \bottomrule
  \end{tabular}
  \vspace{5 pt}
  \caption{Performance of PointPillars and SECOND under the proposed multi-task framework on the nuScenes validation set.}
  \label{tab:otherdetectors}
\end{table}


\textbf{Compatibility with other BEV-based 3D object detection models}
To demonstrate that our proposed multi-task framework can potentially improve the performance of any BEV-based 3D object detection method, we ran another set of experiments, with two different 3D object detection methods, PointPillars \cite{lang2019pointpillars} and SECOND \cite{yan2018second}. While the detection backbone and the detection head are swapped, the rest of the framework and experiment setup are unchanged. As shown in Table \ref{tab:otherdetectors}, when integrated as part of the multi-task framework, the performance of these two detectors are improved significantly. This demonstrates that the effectiveness of our framework is universal across different BEV-based 3D detection methods. More specifically, using a feature weighting mechanism to combine multi-task, multi-view features provides an unintrusive way to enrich any BEV-based detection backbone. Furthermore, feature maps that embed potential locations of object boundaries and centers are well received by any detection head.


\begin{table}[t]
  \centering
  \begin{tabular}{*{3}{c}}
    \toprule
    \textbf{Method} & \textbf{mAP} & \textbf{NDS} \\
    \midrule
    Single-task learning & 59.9 & 66.8 \\
    Multi-task learning & \textbf{60.3} & \textbf{67.1} \\
    \bottomrule
  \end{tabular}
  \vspace{5 pt}
  \caption{Performance comparison between the proposed multi-task learning and the single-task learning (pre-trained CPSeg with frozen weights) on the nuScenes validation set.}
  \label{tab:pretrained}
\end{table}

\textbf{Effects of using a pre-trained CPSeg model}
Another alternative for guiding the 3D object detection model is to pre-train the panoptic segmentation model prior to training the 3D object detection model (Single-task learning). We pre-trained the CPSeg model using panoptic targets, and subsequently trained the CenterPoint model while keeping the weights of CPSeg frozen. From the results in Table \ref{tab:pretrained}, it can be seen that multi-task learning has a superior performance. This shows that when jointly trained, CPSeg learns to pick up RV features that not only benefits the panoptic segmentation task, but also guides the detection backbone.

\subsection{Limitation and Future Work}
Despite observing a boost in performance, integrating the object detection method as part of our multi-task framework has a shortcoming. The proposed framework is composed of two separate backbones, which increases the overall framework complexity. Despite some modifications to simplify both backbones, our proposed method runs at 6 FPS on the nuScenes dataset, below the 10 FPS target that is necessary for the model to operate in real-time. For future works, we plan to design a shared backbone for both 3D panoptic segmentation and object detection for a reduced complexity and faster run-time. This shared backbone should retain the ability of multi-view feature extraction in order to maintain the current high detection accuracy.

\section{Conclusions}
We propose a framework for guiding the LiDAR-based 3D object detection method using panoptic segmentation. In this framework, the RV-based feature maps of the panoptic segmentation model backbone are used to augment the BEV-based feature maps of the detection model. Furthermore, the semantic information estimated by the panoptic segmentation model is used to highlight the location of each class of foreground objects in the detection backbone. Moreover, the instance-level information is used to guide the detection head to attend to possible centers of each object bounding box in the BEV plane. Experimental results on the nuScenes dataset, which provides both the panoptic segmentation and 3D box labels, demonstrate the effectiveness of the proposed framework for increasing the detection accuracy and score of multiple BEV-based 3D detection methods.

\appendix

\bibliographystyle{unsrt}  
\bibliography{references}

\end{document}